\documentclass[10pt,twocolumn,letterpaper]{article}

\usepackage{cvpr}
\usepackage{times}
\usepackage{epsfig}
\usepackage{graphicx}
\usepackage{amsmath}
\usepackage{amssymb}
\usepackage{booktabs}

\usepackage[numbers,sort]{natbib}
\usepackage{xcolor, colortbl}
\usepackage{tabularx}
\usepackage{multirow}

\usepackage{pbox}

\newcolumntype{C}[1]{>{\centering\let\newline\\\arraybackslash\hspace{0pt}}p{#1}}

\definecolor{brown}{rgb}{0.65, 0.16, 0.16}
\definecolor{purp}{rgb}{0.65, 0.16, 0.65}
\definecolor{wine}{rgb}{0.45, 0.18, 0.22}

\DeclareMathOperator*{\argmax}{arg\!\max}

\usepackage[pagebackref=true,breaklinks=true,letterpaper=true,colorlinks,bookmarks=false]{hyperref}

\cvprfinalcopy 

\def\cvprPaperID{3961} 

\ifcvprfinal\fi
\begin{document}

\title{Domain-Specific Batch Normalization for Unsupervised Domain Adaptation}
	
\author{
Woong-Gi Chang\footnotemark[1] ~$^{1, 2}$ \quad Tackgeun You\footnotemark[1] ~$^{1, 2}$  \quad  Seonguk Seo\footnotemark[1]\thanks{indicates equal contribution.} ~$^1$ \quad Suha Kwak$^{2}$ \quad Bohyung Han$^1$ \\
 $^1$Computer Vision Lab., ECE \& ASRI, Seoul National University, Korea\\
 $^2$Computer Vision Lab., CSE, POSTECH, Korea\\
}

\maketitle


\begin{abstract}

We propose a novel unsupervised domain adaptation framework based on domain-specific batch normalization in deep neural networks.
We aim to adapt to both domains by specializing batch normalization layers in convolutional neural networks while allowing them to share all other model parameters, which is realized by a two-stage algorithm.
In the first stage, we estimate pseudo-labels for the examples in the target domain using an external unsupervised domain adaptation algorithm---for example, MSTN~\cite{MSTN} or CPUA~\cite{CPUA}---integrating the proposed domain-specific batch normalization.
The second stage learns the final models using a multi-task classification loss for the source and target domains.
Note that the two domains have separate batch normalization layers in both stages. 
Our framework can be easily incorporated into the domain adaptation techniques based on deep neural networks with batch normalization layers.
We also present that our approach can be extended to the problem with multiple source domains. 
The proposed algorithm is evaluated on multiple benchmark datasets and achieves the state-of-the-art accuracy in the standard setting and the multi-source domain adaption scenario.
\end{abstract}


\section{Introduction}
\label{sec:introduction}
Unsupervised domain adaptation is a learning framework to transfer knowledge learned from source domains with a large number of annotated training examples to target domains with unlabeled data only.
This task is challenging due to domain shift problem, which is the phenomenon that source and target datasets have different characteristics.
The domain shift is common in real-world problems and should be handled carefully for broad applications of trained models.
Unsupervised domain adaptation aims to learn robust models for handling the issue, and is getting popular these days since it can be a rescue for visual recognition tasks relying on the datasets with limited diversity and variety. 

Recent advances in unsupervised domain adaptation owe to its success of deep neural networks. 
Traditional domain adaptation techniques based on shallow learning are reformulated using deep neural networks with proper loss functions. 
Strong representation power of deep networks rediscovers effectiveness of the previous methods and facilitates development of brand-new algorithms.
There is a large volume of research on unsupervised domain adaptation based on deep neural networks~\cite{DANN,MSTN,CPUA,iCAN,CDAN,hoffman2017cycada,DeepCORAL}, and in recent years we have witnessed significant performance improvement.

One of the drawbacks in many existing unsupervised domain adaptation techniques~\cite{DANN,MSTN,CPUA,hoffman2017cycada,DeepCORAL} is the fact that source and target domains share the whole network for training and prediction.
Shared components between both domains are inevitable because there is something common in the two domains; we often need to rely on information in the source domain to learn the networks adapting to unlabeled target domain data. 
However, we believe that better generalization performance can be achieved by separating domain-specific information from domain-invariant one since the two domains obviously have different characteristics that are not compatible within a single model.

To separate domain-specific information for unsupervised domain adaptation, we propose a novel building block for deep neural networks, referred to as \emph{Domain-Specific Batch Normalization} (DSBN). 
A DSBN layer consists of two branches of Batch Normalization (BN), each of which is in charge of a single domain exclusively.
DSBN captures domain-specific information using BN parameters and transforms domain-specific data into domain-invariant representations using the parameters. 
Since this idea is generic, DSBN is universally applicable to various deep neural networks for unsupervised domain adaptation that have BN layers. 
Furthermore, it can be easily extended for multi-source domain adaptation scenarios.

Based on the main idea, we introduce a two-stage framework based on DSBN for unsupervised domain adaptation, where our network first generates pseudo-labels of unlabeled data in the target domain and then learns a fully supervised model using the pseudo-labels. 
Specifically, the first stage estimates the initial pseudo-labels of target domain data through an existing unsupervised domain adaptation network incorporating DSBN.
In the second stage, a multi-task classification network with DSBN layers is trained with full supervision using the data from both source and target domains, where the pseudo-labels generated at the first stage are assigned to the target domain data. 
To further improve accuracy, we iterate the second stage training and refine the labels of the examples in the target domain.

Our main contributions are summarized as follows:
\begin{itemize}
	\item We propose a novel unsupervised domain adaptation framework based on DSBN, which is a generic method applicable to various deep neural network models for domain adaptation.
	\item We introduce a two-stage learning method with DSBN comprising pseudo-label estimation followed by multi-task classification, which is naturally integrated into existing unsupervised domain adaptation methods.
	\item Our framework provides a principled algorithm for unsupervised domain adaptation with multiple sources via its straightforward extension.
	\item We achieve the state-of-the-art performance on the standard benchmarks including Office-31 and VisDA-C datasets by integrating our framework with two recent domain adaptation techniques.
\end{itemize}

The rest of our paper has the following organization.
We first review unsupervised domain adaptation approaches in Section~\ref{sec:related}, and then discuss two recent backbone models to integrate DSBN in Section~\ref{sec:overview_and_background}.
Section~\ref{sec:dsbn} presents the main idea of DSBN and Section~\ref{sec:two_stage} describes how to integrate our framework into the existing unsupervised domain adaptation techniques. 
We show experimental results in Section~\ref{sec:experiments}, and conclude this paper with a brief remark in Section~\ref{sec:conclusion}.


\section{Related Work}
\label{sec:related}

This section reviews mainstream approaches in unsupervised domain adaptation that are related to our approach.

\subsection{Domain-invariant Learning}
Learning domain-invariant representation is critical for success of unsupervised domain adaptation, and many existing approaches attempt to align data distributions between source and target domains either globally or locally.

Global alignment techniques include Maximum Mean Discrepancy (MMD)~\cite{MMD, WMMD}, Central Moment Discrepancy (CMD)~\cite{CMD}, Wassterstein distance~\cite{shen2017adversarial}, and CORrelation ALignment (CORAL)~\cite{CORAL},
some of which are reformulated using deep neural networks~\cite{JAN, WMMD, DeepCORAL}.
Also, adversarial learning has been widely used for global domain alignment.
Domain Adversarial Neural Networks (DANN)~\cite{DANN} learn domain-invariant representations using domain adversarial loss while Adversarial Discriminative Domain Adaptation (ADDA)~\cite{ADDA} combines adversarial learning with discriminative feature learning.
Joint Adaptation Network (JAN)~\cite{JAN} combines adversarial learning with MMD~\cite{MMD} by aligning the joint distributions of multiple domain-specific layers across domains.

Local alignment approaches learn domain-invariant models by aligning examples of the same classes across source and target domains.
Cycle-Consistent Adversarial Domain Adaptation (CyCADA)~\cite{hoffman2017cycada} introduces a cycle-consistency loss, which enforces the model to ensure a correct semantic mapping.
Conditional Domain Adaptation Network (CDAN)~\cite{CDAN} conditions an adversarial adaptation model on discriminative information conveyed in the classifier predictions.

\subsection{Domain-specific Learning}
Approaches in this category learn domain-specific information together with domain-invariant one for domain adaptation.
Domain Separation Network (DSN)~\cite{DSN} learns feature representations by separating domain-specific network components from shared ones.
Collaborative and Adversarial Network (CAN)~\cite{iCAN} proposes a new loss function that enforces lower-level layers of the network to have high domain-specificity and low domain-invariance while making higher-level layers to have the opposite properties.

Batch normalization parameters have been used to model domain-specific information in unsupervised domain adaptation scenarios.
AdaBN~\cite{adaptiveBN} proposes a post-processing method to re-estimate batch normalization statistics using target samples.
Domain alignment layers used in~\cite{autoDIAL} deal with the domain shift problem by aligning source and target distributions to a reference Gaussian distribution.
Also, \cite{mancini2018boosting} learns to automatically discover multiple latent source domains, which are used to align target and source distributions.
Note that our algorithm is more flexible and efficient as it learns domain-specific properties only with affine parameters of batch normalization and all network parameters except them are used for domain invariant representation.

\subsection{Learning with Pseudo Labels}
Recent algorithms often estimate pseudo labels in target domain and directly learn a domain-specific model for target domain.
To obtain the pseudo labels, Moving Semantic Transfer Network (MSTN)~\cite{MSTN} exploits semantic matching and domain adversarial losses while Class Prediction Uncertainty Alignment (CPUA)~\cite{CPUA} employs class scores as feature for adversarial learning.
Asymmetric Tri-training Network (ATN)~\cite{ATN} and Collaborative and Adversarial Network (CAN)~\cite{iCAN} estimate pseudo labels based on the consensus of multiple models.

Our framework also exploits pseudo labels to learn domain-specific models effectively. 
Compared to the existing method, our framework estimates more reliable pseudo labels since domain-specific information is captured effectively by domain-specific batch normalization.


\section{Preliminaries}
\label{sec:overview_and_background}
In unsupervised domain adaptation, we are given two datasets;
$\mathcal{X}_S$ is for the labeled source domain and  $\mathcal{X}_T$ is the unlabeled dataset for target domain,
where $n_S$ and $n_T$ denote the cardinalities of $\mathcal{X}_S$ and $\mathcal{X}_T$, respectively. 
Our goal is to classify examples in the target domain by transferring knowledge of classification learned from the source domain based on full supervision.
This section discusses the details of two state-of-the-art approaches for the integration of our domain-specific batch normalization technique.

\subsection{Moving Semantic Transfer Network}
\label{sub:mstn}

MSTN~\cite{MSTN} proposes a semantic matching loss function to align the centroids of the same classes across domains, based on pseudo-labels of unlabeled target domain samples.
The formal definition of the total loss function $\mathcal{L}$ is given by
\begin{equation}
\mathcal{L} = \mathcal{L}_\text{cls}(\mathcal{X}_S) + \lambda \mathcal{L}_\text{da}(\mathcal{X}_S, \mathcal{X}_T) + \lambda\mathcal{L}_\text{sm}(\mathcal{X}_S, \mathcal{X}_T).
\label{eq:mstn_loss}
\end{equation}
The classification loss $\mathcal{L}_\text{cls} (\mathcal{X}_S)$ is the cross-entropy loss for source dataset, and the domain adversarial loss $\mathcal{L}_\text{da}$ makes a network confused about the domain membership of an example as discussed in \cite{DANN}.
The semantic matching loss aligns the centroids of the same classes across domains. 
Note that pseudo-labels should be estimated to compute the semantic matching losses.
Intuitively, the loss function in Eq.~\eqref{eq:mstn_loss} encourages two domains to have the same distribution, especially by adding adversarial and semantic matching loss terms. 
Hence, the learned network based on the loss function can be applied to examples in target domain.


\subsection{Class Prediction Uncertainty Alignment }
\label{sub:cpua}
CPUA~\cite{CPUA} is a strikingly simple approach that only aligns the class probabilities across domains.
CPUA addresses class imbalance issue in both domain and introduces a class weighted loss function to exploit the class priors.

Let $p_S(c) = {n_S^c}/{n_S}$ be the fraction of source samples that have class label $c$ and $\widetilde{p}_T(c) = n_T^{c}/ n_T$ be the fraction of targets samples that have pseudo-label $c$. $n_T^{c}$  denotes the cardinality of $\{x \in \mathcal{X}_T \mid \widetilde{y}(x) = c\}$, where $\widetilde{y}(x) = \argmax_{i\in C} F(x)[i]$.
Then the class weights for each domain is respectively given by
\begin{align}
  \label{eq:weight_src}
  w_S(x,y) &= \frac{\max_{y'} p_S(y') }{ p_S(y) } \\
  &\hspace{-3.5cm} \text{and} \nonumber \\
  w_T(x) &= \frac{\max_{y'} \widetilde{p}_T(y') }{ \widetilde{p}_T(\widetilde{y}(x)) }.
\end{align}
Their total loss function can be written as
\begin{equation}
\label{eq:cpua_loss}
\mathcal{L} = \mathcal{L}_\text{cls}(\mathcal{X}_S) + \lambda \mathcal{L}_\text{da}(\mathcal{X}_S, \mathcal{X}_T),
\end{equation}
where
\begin{align}
\mathcal{L}_\text{cls}(\mathcal{X}_S) 
=& \frac{1}{n_S} \sum_{(x,y)\in\mathcal{X}_S}{w_S(x,y) \ell(F(x), y)}, \\
  \mathcal{L}_\text{da}(\mathcal{X}_S, \mathcal{X}_T) =& \frac{1}{n_S} \sum_{(x,y) \in \mathcal{X}_S} w_S(x,y) \ell(D(F(x))), 1) \nonumber \\
  +& \frac{1}{n_T} \sum_{x \in \mathcal{X}_T} w_T(x) \ell(D(F(x))), 0).
\end{align}
Note that $F(\cdot)$ is a classification network, $\ell(\cdot, \cdot)$ denotes the cross-entropy loss and $D(\cdot)$ is a domain discriminator.

\begin{figure*}
   \begin{center}
      \includegraphics[width=\linewidth]{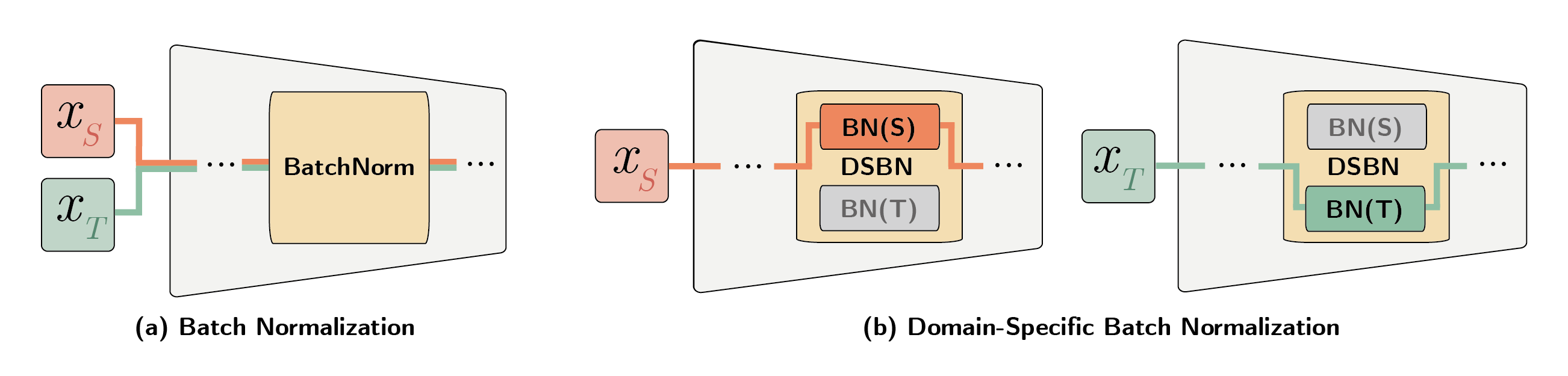}
   \end{center}
   \vspace{-0.6cm}
   \caption{Illustration of difference between BN and DSBN. A DSBN layer consists of two branches in the batch normalization layer---one is for source domain ($S$), and the other is for target domain ($T$). 
Each input example chooses one of the branches according to its domain. 
In a domain adaptation network with DSBN layers, all parameters except those of DSBN are shared across the two domains and learn the information common in both domains effectively, while domain-specific properties are captured efficiently through the domain specific BN parameters of DSBN layers. 
Note that DSBN layers can be plugged in any unsupervised domain adapatation networks with BN layers.}
   \label{fig:bndsbn}
\end{figure*}


\section{Domain-Specific Batch Normalization}
\label{sec:dsbn}
This section briefly reviews Batch Normalization (BN) for a comparison to our DSBN, and then presents DSBN and its extension for multi-source domain adaptation.

\subsection{Batch Normalization}
\label{sub:bn}
BN~\cite{Batchnorm} is a widely used training technique in deep networks. 
A BN layer whitens activations within a mini-batch of $N$ examples for each channel dimension and transforms the whitened activations using affine parameters $\gamma$ and $\beta$.
Denoting by $\mathbf{x} \in \mathbb{R}^{H\times W\times N}$ activations in each channel, BN is expressed as
\begin{align}
\text{BN}(\mathbf{x}[i, j, n] ; \gamma, \beta) &= \gamma \cdot \hat{\mathbf{x}} [i, j, n] + \beta,
\end{align}
where
\begin{align}
\hat{\mathbf{x}}[i, j, n] &=  \frac{\mathbf{x}[i, j, n] - \mu}{\sqrt{\sigma^2 + \epsilon}}.
\end{align}
The mean and variance of activations within a mini-batch, $\mu$ and $\sigma$, are computed by
\begin{align}
\mu &= \frac{\sum_n \sum_{i,j} \mathbf{x}[i,j, n]}{N \cdot H \cdot W}, \\
\sigma^2 &= \frac{\sum_n \sum_{i,j} \left( \mathbf{x}[i, j, n] - \mu \right)^2 }{N \cdot H \cdot W}.
\end{align}
and $\epsilon$ is a small constant to avoid divide-by-zero.

During training, BN estimates the mean and variance of the entire activations, denoted by $\bar{\mu}$ and $\bar{\sigma}$, through exponential moving average with update factor $\alpha$. 
Formally, given the $t^\text{th}$ mini-batch, the mean and variance are given by
\begin{align}
\bar{\mu}^{t+1} &= (1-\alpha)\bar{\mu}^{t} + \alpha\mu^{t}, \\	
\left(\bar{\sigma}^{t+1}\right)^2 &= (1-\alpha) \left( \bar{\sigma}^{t} \right)^2 + \alpha \left( \sigma^t \right)^2.
\end{align}
In the testing phase, BN uses the estimated mean and variance for whitening input activations.
Note that sharing the mean and variance for both source and target domain are inappropriate if domain shift is significant.

\subsection{Domain-Specific Batch Normalization}
\label{sub:dsbn_intg}
DSBN is implemented by using multiple sets of BN~\cite{Batchnorm} reserved for each domain.
Figure~\ref{fig:bndsbn} illustrates the difference between BN and DSBN. 
Formally, DSBN allocates domain-specific affine parameters $\gamma_d$ and $\beta_d$ for each domain label $d \in \{S, T\}$.
Let $\mathbf{x}_d \in \mathbb{R}^{H\times W\times N}$ denote activations at each channel belong to a domain label $d$, then DSBN layer can be written as
\begin{align}
\text{DSBN}_d(\mathbf{x}_d[i, j, n] ; \gamma_d, \beta_d) &= \gamma_d \cdot \hat{\mathbf{x}}_d [i, j, n] + \beta_d,
\end{align}
where
\begin{align}
\hat{\mathbf{x}}_d[i, j, n] =  \frac{\mathbf{x}_d[i, j, n] - \mu_d}{\sqrt{\sigma_{d}^{2} + \epsilon}}, 
\end{align}
and
\begin{align}
\mu_d &= \frac{\sum_n \sum_{i,j} \mathbf{x}_d[i,j, n]}{N \cdot H \cdot W}, \\
\sigma_{d}^2 &= \frac{\sum_n \sum_{i,j} \left( \mathbf{x}_d[i, j, n] - \mu_d \right)^2 }{N \cdot H \cdot W}.
\end{align}

During training, DSBN estimates the mean and variance of activations for each domain separately by exponential moving average with update factor $\alpha$, which is given by 
\begin{align}
\bar{\mu}_d^{t+1} &= (1-\alpha) \bar{\mu}_d^{t} + \alpha \mu_d^t , \\	
\left(\bar{\sigma}_d^{t+1}\right)^2 &= (1-\alpha) \left(\bar{\sigma}_d^{t}\right)^2 + \alpha \left( \sigma_d^t \right)^2 .
\end{align}
The estimated mean and variance for each domain are used for the examples in the corresponding domain at the testing phase of DSBN.

We expect DSBN to capture the domain-specific information by estimating batch statistics and learning affine parameters for each domain separately.
We believe that DSBN allows the network to learn domain-invariant features better because domain-specific information within the network can be removed effectively by exploiting the captured statistics and learned parameters from the given domain.

DSBN is easy to be plugged in existing deep neural networks for unsupervised domain adaptation.
An existing classification network $F(\cdot)$ can be converted to a domain-specific network by replacing all BN layers with DSBN layers and training the entire network using data with domain labels.
The domain-specific network is denoted by $F_d(\cdot)$, which is specialized to either the source or the target domain depending on the domain variable $d\in \{ S, T \}$.

\subsection{Extension to Multi-Source Domain Adaptation}
DSBN is easily extended for multi-source unsupervised domain adaptation by adding more domain branches in it.
In addition, a new loss function for the multi-source domain adaptation is simply defined by the sum of losses from all source domains as follows:
\begin{align}
& \mathcal{L} = \frac{1}{ |{\mathcal{D}_S}| }\sum_{i}^{|{\mathcal{D}_S}|}  \Big( \mathcal{L}_\text{cls}(\mathcal{X}_{S_i}) + \mathcal{L}_\text{align}(\mathcal{X}_{S_i}, \mathcal{X}_T) \Big),
\end{align}
where $\mathcal{D}_S = \{\mathcal{X}_{S_1}, \mathcal{X}_{S_2}, ...\}$ is a set of source domains and $\mathcal{L}_\text{align}$ can be any kind of loss for aligning source and target domains.
The remaining procedure for training is identical to the single-source domain adaptation case.


\begin{figure}[t]
	\begin{center}
		\includegraphics[width=\linewidth]{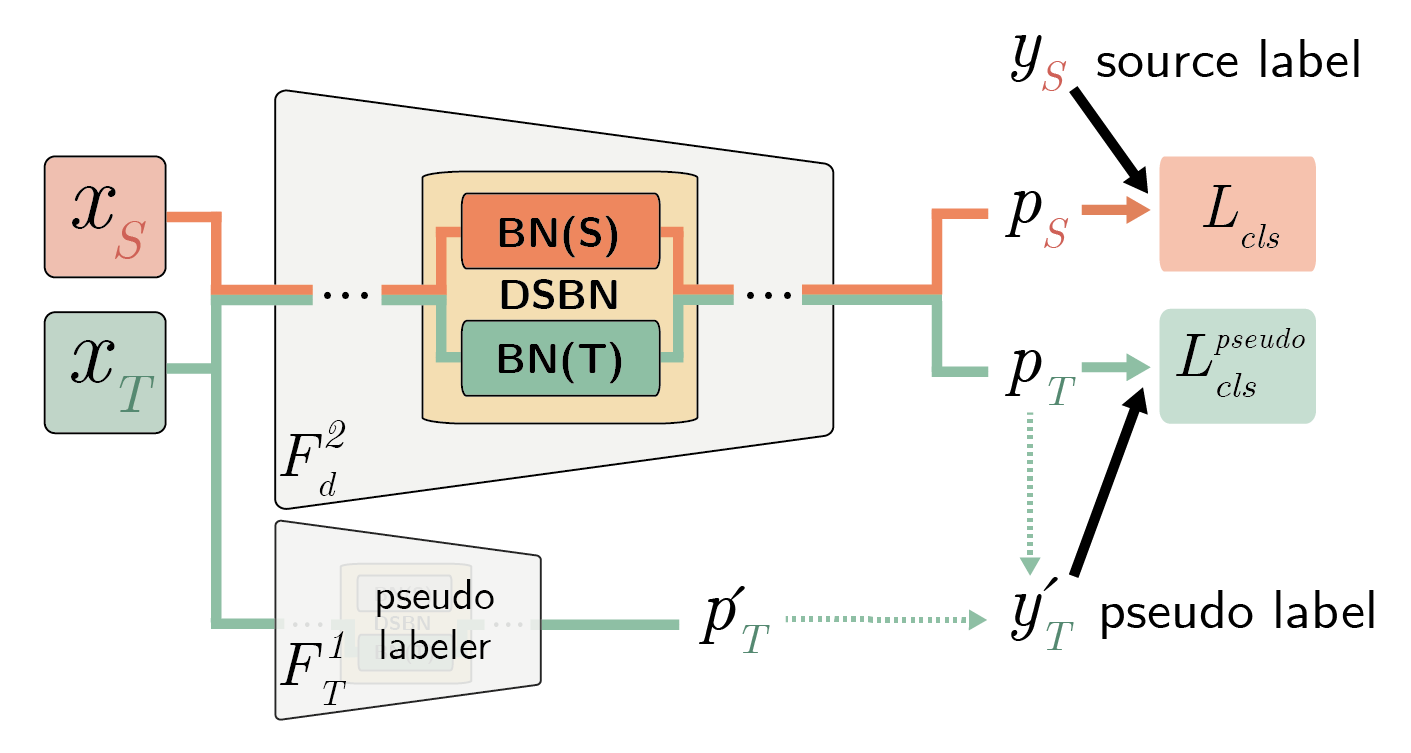}
	\end{center}
	\vspace{-0.3cm}
	\caption{Overview of the second stage training.
		To use intermediate pseudo labels for target domain samples, we employ the network trained in the first stage,  $F^1_T(x)$, as a pseudo labeler in the second stage.
		In this stage, the network is trained with classification losses only in both domains.}
	\label{fig:overview}
\end{figure}

\begin{figure*}
	\begin{center}
		\includegraphics[width=0.9\linewidth]{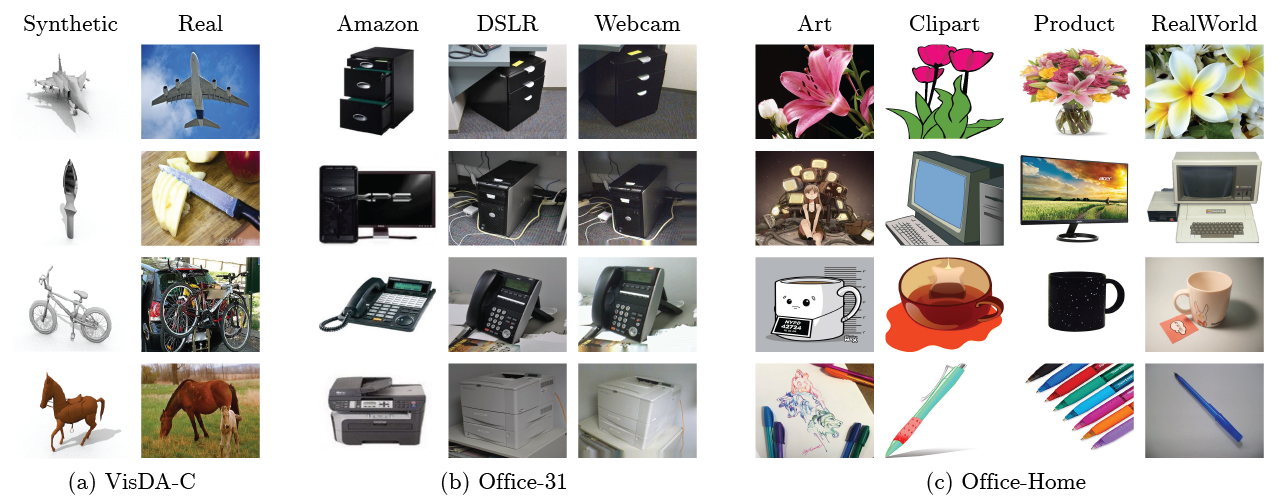}
	\end{center}
	\vspace{-0.3cm}
	\caption{
	Sample images of each dataset. (a) VisDA-C dataset images of two domains, (b) Office-31 dataset images of three domains, (c) Office-Home dataset images of four domains.
	}
	\label{fig:dataset}
\end{figure*}

\section{Domain Adaptation with DSBN}
\label{sec:two_stage}

DSBN is a generic technique for unsupervised domain adaptation, and can be integrated into various algorithms based on deep neural networks with batch normalization.
Our framework trains deep networks for unsupervised domain adaptation in two stages.
In the first stage, we train an existing unsupervised domain adaptation network to generate initial pseudo-labels of target domain data.
The second stage learns the final models of both domains using the ground-truth in the source domain the pseudo-labels in the target domain as supervision, where the pseudo-labels in the target domain are refined progressively during the training procedure.
The networks in both stages incorporate DSBN layers to learn domain-invariant representations more effectively and better adapt to the target domain consequently. 
To further improve accuracy, we can perform additional iterations of the second stage training, where the pseudo-labels are updated using the results in the preceding iteration.
The remaining parts of this section present details of our two stage training method with DSBN.

\subsection{Stage 1: Training Initial Pseudo Labeler}
\label{sub:pseudo_labeler}
Since our framework is generic and flexible, any unsupervised domain adaptation network can be employed to estimate initial pseudo-labels of target domain data as long as it has BN layers.
In this paper, we choose two state-of-the-art models as the initial pseudo-label generator: MSTN~\cite{MSTN} and CPUA~\cite{CPUA}.
As described in Section~\ref{sub:dsbn_intg}, we replace their BN layers with DSBNs so that they learn domain-invariant representations more effectively.
The networks are then trained by their original losses and learning strategies.
A trained initial pseudo-label generator is denoted by $F^1_T$.

\subsection{Stage 2: Self-training with Pseudo Labels}
\label{sub:refinement}

In the second stage, we utilize data and their labels from both domains to take advantage of rich domain-invariant representations and train the final models for both domains with full supervision.
The network is trained with two classification losses---one for the source domain with ground-truth labels and the other for the target domain with pseudo-labels---and the resulting networks are denoted by $F^2_d$ where $d\in \{ S,T\}$. 
The total loss function is given by a simple summation of two loss terms from two domains as follows:

\begin{align}
\mathcal{L} &= \mathcal{L}_\text{cls}(\mathcal{X}_S) + \mathcal{L}_{\text{cls}}^{\text{pseudo}}(\mathcal{X}_T)
\end{align}
where
\begin{align}
\mathcal{L}_\text{cls}(\mathcal{X}_S) &= \hspace{-0.3cm} \sum_{(x, y)\in\mathcal{X}_S}{\ell ({F^2_S(x), y})}, \label{eq:l_cls_src} \\
\mathcal{L}_\text{cls}^\text{pseudo}(\mathcal{X}_T) &= \sum_{x\in\mathcal{X}_T}{\ell ({F^2_T(x), y'})}. \label{eq:l_cls_tgt}
\end{align}
In Eq.~\eqref{eq:l_cls_src}~and~\eqref{eq:l_cls_tgt}, $\ell(\cdot, \cdot)$ is the cross-entropy loss and $y'$ denotes the pseudo-label assigned to a target domain example $x \in \mathcal{X}_T$.
	
The pseudo-label $y'$ is initialized by $F^1_T$ and progressively refined by $F^2_T$ as follows:
\begin{equation}
y^{\prime} = \argmax_{c \in C} \Big\{ (1-\lambda) F_T^1(x)[c] + \lambda F_T^2 (x)[c] \Big\},
\label{eq:scheduled_ensemble}
\end{equation}
where $F_T^i(x)[c]$ indicates the prediction score of the class $c$ given by $F_T^i$ and $\lambda$ is a weight factor that changes gradually from 0 to 1 during training.
This approach can be considered as a kind of self-training since $F_T^2$ takes part in the pseudo-label generation while trained.
At an early phase of training, we put more weights on the initial pseudo-labels given by $F_T^1$ since the prediction by $F_T^2$ may be unreliable.
The weight $\lambda$ then increases gradually, and at the last phase of training, the pseudo labels rely totally on $F_T^2$.
We follow \cite{DANN} to suppress potentially noisy pseudo-labels; the adaptation factor $\lambda$ is gradually increased by $\lambda=\frac{2}{1+\text{exp}(-\gamma \cdot p)}-1$ with $\gamma = 10$.

Target domain images are recognized more accurately by $F^2_T$ than $F^1_T$ in general since $F^2_T$ exploits reasonable initial pseudo-labels given by $F^1_T$ for training while $F^1_T$ relies only on weak information for domain alignment.
It is natural to employ $F^2_T$ to estimate more accurate initial pseudo-labels for the purpose of further accuracy improvement.
Thus, we conduct the second stage procedure iteratively, where the initial pseudo-labels are updated using the prediction results of the model in the preceding iteration.
We emperically observe that this iterative approach is effective to improve classification accuracy in the target domain.

\begin{table*}[!t]
	\centering
	\caption{Classification performance (\%) of multiple algorithms on VisDA-C validation dataset using a ResNet-101 backbone network. The results clearly show that our two-stage learning framework with DSBN is effective to improve accuracy. }
	\label{tab:visda_result}
	\vspace{0.1cm}
	\scalebox{0.9}{
	\begin{tabular}{llllllllllllll}  
		\toprule
		{Method} &  {aero} & bicycle & bus & car & horse & knife & {motor} & person & plant & {skate} & train & truck & Avg \\ \hline
		Source only		&55.1&53.3&61.9&59.1&80.6&17.9&79.7&31.2&81.0&26.5&73.5&8.5&52.4\\
		DAN~\cite{DAN}	&87.1&63.0&76.5&42.0&90.3&42.9&85.9&53.1&49.7&36.3&85.8&20.7&61.1\\
		DANN~\cite{DANN}	&81.9&77.7&82.8&44.3&81.2&29.5&65.1&28.6&51.9&54.6&82.8&7.8&57.4\\
		MCD~\cite{saito2017maximum}		&87.0&60.9&83.7&64.0&88.9&\bf{79.6}&84.7&76.9&88.6&40.3&83.0&25.8&71.9\\
		ADR~\cite{saito2017adversarial}	&87.8&79.5&83.7&65.3&92.3&61.8&88.9&73.2&87.8&60&85.5&32.3&74.8\\
		\hline
		MSTN (reproduced) 	&89.3&49.5&74.3&67.6&90.1&16.6&93.6&70.1&86.5&40.4&83.2&18.5&65.0\\
		~~with DSBN (Stage 1) &90.3&78.4&75.0&53.5&90.1&42.8&85.5&76.5&86.7&64.1&81.5&43.6&72.3 \\
		~~with DSBN (Stage 1 and 2) &\bf{94.7}&\bf{86.7}&76.0&\bf{72.0}&\bf{95.2}&75.1&87.9&\bf{81.3}&\bf{91.1}&68.9&\bf{88.3}&45.5&\bf{80.2}\\
		\hline
		CPUA (reproduced)	&90.7&53.4&79.6&59.9&87.9&18.7&\bf{94.5}&61.6&90.4&51.3&81.8&29.3&66.6\\
		~~with DSBN (Stage 1) &91.1&77.2&\bf{84.0}&49.5&90.7&37.9&87.8&76.6&85.6&60.9&78.1&43.9&71.9\\
		~~with DSBN (Stage 1 and 2) &93.0&83.6&79.9&55.8&94.7&20.6&89.7&80.2&\bf{91.1}&\bf{80.8}&84.8&\bf{59.7}&76.2 \\
		\bottomrule
	\end{tabular}
	}
\vspace{-0.2cm}
\end{table*}

\begin{table*}[t]
\centering
\caption {Classification accuracies (\%) on Office-31 dataset (ResNet-50). *The original paper reported average accuracy of 79.1\% with AlexNet.$^\dagger$The original paper reported average accuracy of 87.9\% with ResNet-50.} 
\label{tab:officefulltable}
\vspace{0.1cm}
	\scalebox{0.9}{
	
\begin{tabular}{lllllllll}
\toprule

{Method} & A $\rightarrow$ W & W $\rightarrow A$ & A $\rightarrow$ D & D $\rightarrow$ A & W $\rightarrow$ D & D $\rightarrow$ W & Avg & \\ \hline
Source Only    & 73.5 & 59.8 & 76.5 & 56.7 & 99.0 & 93.6 & 76.5 \\
DDC~\cite{DDC}         & 76.0 & 63.7 & 77.5 & 67.0 & 98.2 & 94.8 & 79.5 \\
DAN~\cite{DAN}         & 80.5 & 62.8 & 78.6 & 63.6 & 99.6 & 97.1 & 80.4 \\
RTN~\cite{RTN}         & 84.5 & 64.8 & 77.5 & 66.2 & 99.4 & 96.8 & 81.6 \\
DANN~\cite{DANN}          & 79.3 & 63.2 & 80.7 & 65.3 & 99.6 & 97.3 & 80.9\\
JAN~\cite{JAN}         & 86.0 & 70.7 & 85.1 & 69.2 & 99.7 & 96.7 & 84.6 \\
iCAN~\cite{iCAN}        & 92.5 & 69.9 & 90.1 & 72.1 & \bf{100.0}& 98.8 & 87.2 \\
CDAN-M~\cite{CDAN}& 93.1 &  70.3 & \bf{93.4} & 71.0& \bf{100.0} & 98.6 & 87.7 \\
\hline
MSTN (reproduced)~\cite{MSTN}$^*$& 91.3& 65.6& 90.4 & \bf{72.7} & \bf{100.0} & 98.9 & 86.5 \\
~~with DSBN (Stage 1) & {92.5} & {73.1} & {90.6} & 72.2 & \bf{100.0} & 98.5 & {87.8} \\
~~with DSBN (Stage 1 and 2) & 92.7& \bf{74.4}&92.2&71.7&\bf{100.0}&99.0&\bf{88.3}\\
\hline
CPUA (reproduced)~\cite{CPUA}$^\dagger$ &90.1 & 71.6 & 86.8 & 71.3 & \bf{100.0} & 98.6 & 86.4 \\
~~with DSBN (Stage 1) & {92.3} & {72.7} & {88.8} & {72.0} & \bf{100.0} &  \bf{99.1} & {87.5}  \\		
~~with DSBN (Stage 1 and 2) & \bf{93.3} & 73.9 & 90.8 & \bf{72.7} & \bf{100.0} & \bf{99.1} & \bf{88.3} \\
\bottomrule
\end{tabular}}
\end{table*}

\section{Experiments}
\label{sec:experiments}
We present the empirical results to validate the proposed framework and compare our method with the state-of-the-art domain adaptation methods.

\subsection{Experimental Settings}
We discuss datasets used for training and evaluation and present implementation details including hyperparameter setting.

\vspace{-0.2cm}
\paragraph{Datasets}
We employ three datasets for our experiment: VisDA-C~\cite{visda}, Office-31~\cite{office-31} and Office-Home~\cite{office-home}.
VisDA-C is a large-scale benchmark dataset used for Visual Domain Adaptation Challenge 2017.
It consists of two domains---\emph{synthetic} and \emph{real}---and have 152,409 synthetic images and 55,400 real images of 12 common object classes from MS-COCO~\cite{Mscoco} dataset.
Office-31 is a standard benchmark for domain adaptation, which consists of three different domains in 31 categories: Amazon (A) with 2,817 images, Webcam (W) with 795 images and DSLR (D) with 498 images.
Office-Home~\cite{office-home} has four domains: Art (Ar) with 2,427 images, Clipart (Cl) with 4,365 images, Product (Pr) with 4,439 images, and Real-World (Rw) with 4,357 images.
Each domain contains 65 categories of common daily objects. We adopt the fully transductive protocol introduced in~\cite{DANN} to evaluate our framework on the datasets.

\vspace{-0.2cm}
\paragraph{Implementation details}
Following~\cite{DANN, saito2017maximum}, as the backbone networks of our framework, we adopt ResNet-101 for the VisDA-C dataset and ResNet-50 for Office-31 and Office-Home datasets. 
All the networks have BN layers and are pre-trained on the ImageNet.
To compare the sheer difference between BN and DSBN layers, we construct mini-batches for each domain and forward them separately.
The batch size is set to 40, which is identical for all experiments.
We use Adam optimizer~\cite{Adamsolver} with $\beta_1=0.9, \beta_2=0.999$. We set initial learning rate $\eta_0= 1.0\times10^{-4}$ and  $5.0 \times 10^{-5}$ for stage 1 and 2, respectively.
As suggested in \cite{DANN}, the learning rate is adjusted by a formula, $\eta_p= \frac{\eta_0}{(1+\alpha p)^{\beta}}$, where $\alpha=10$, $\beta=0.75$, and $p$ denotes training progress linearly changing from 0 to 1.
The maximum number of iterations of the optimizer is set to 50,000.

\begin{table}[t]
\caption{Classification accuracies (\%) in the multi-source scenario on Office-31 dataset using MSTN as a baseline. (ResNet-50)}
\label{tab:multi_office_result}
\vspace{0.1cm}
\scalebox{0.9}{
\setlength{\tabcolsep}{5pt}
\hspace{-0.2cm}
\begin{tabular}{lcccccc}
\toprule
\textit{Single} & A$\rightarrow$W & D$\rightarrow$W & A$\rightarrow$D & W$\rightarrow$D & W$\rightarrow$A & D$\rightarrow$A \\
\hline
BN      & 91.3 & 98.9 & 90.4 & 100.0 & 65.6 & 72.7 \\
DSBN     & 92.5 & 98.5 & 90.6 & 100.0 & 73.1 & 72.2 \\
\toprule
\textit{Merged}        & \multicolumn{2}{c}{(A+D)$\rightarrow$W}& \multicolumn{2}{c}{(A+W) $\rightarrow$D} & \multicolumn{2}{c}{(W+D)$\rightarrow$A}\\
\hline
BN       & \multicolumn{2}{c}{99.6}& \multicolumn{2}{c}{99.6}& \multicolumn{2}{c}{71.3}\\
DSBN     & \multicolumn{2}{c}{99.1}& \multicolumn{2}{c}{99.6}&\multicolumn{2}{c}{73.2}\\
\hline
\textit{Separate}     & \multicolumn{2}{c}{A, D$\rightarrow$W} & \multicolumn{2}{c}{A, W $\rightarrow$ D} & \multicolumn{2}{c}{W, D$\rightarrow$A} \\
\hline
BN       & \multicolumn{2}{c}{\textbf{99.9}} & \multicolumn{2}{c}{99.8} & \multicolumn{2}{c}{69.9} \\
DSBN     & \multicolumn{2}{c}{\textbf{99.9}}& \multicolumn{2}{c}{\textbf{100.0}} & \multicolumn{2}{c}{\textbf{75.6}} \\
\bottomrule
\end{tabular}
}
\end{table}


\subsection{Results}
We present the experimental results on the standard benchmark datasets based on single- and multi-source domain adaptation. 

\vspace{-0.2cm}
\paragraph{VisDA-C} 

Table~\ref{tab:visda_result} quantifies performance of our approach employing MSTN and CPUA as its initial pseudo-label generators, and compares them with state-of-the-art records on the VisDA-C dataset.
In the table, ``$\text{DSBN (Stage 1)}$" means that we replace BN layers with DSBNs and perform the first stage training, and ``$\text{DSBN (Stage 1 and 2)}$" indicates that we conduct both the first and second stage training.
Our proposed method improves accuracy substantially and consistently by applying DSBNs to the baseline models, and achieves the state-of-the-art performance when combined with MSTN~\cite{MSTN}.
Note also that our model reliably recognizes hard classes such as knife, person, skate, and truck.

\vspace{-0.2cm}
\paragraph{Office-31}  
Table \ref{tab:officefulltable} presents overall scores of our approach using MSTN and CPUA on the Office-31 dataset.
Models with DSBN trained in both stages achieve state-of-the-art performance and consistently outperform two baseline models. 
Table \ref{tab:officefulltable} also shows that our framework can be applied to current domain adaptation algorithm successfully and greatly improves performance.

\vspace{-0.2cm}
\paragraph{Multiple Source Domains}  Table~\ref{tab:multi_office_result} and Table~\ref{tab:multi_office_home_result} presents the results of multiple source domain adaptation on the Office-31 and Office-Home datasets, respectively.
To compare multi-source to single-source domain adaptation, we report single-source results on the top of the table as ``{\it Single}" and append two different multi-source scenarios: ``{\it Merged}" and ``{\it Separate}". {\it Merged} means that data from multiple source domains are combined and constructed a new larger source domain dataset while {\it separate} indicates that each source domain is considered separately.
In the {\it separate} case, we have a total of $|\mathcal{D}_\mathcal{S}|+1$ domains and the same number of DSBN branches in the network.
While there is a marginal performance gain between BN and DSBN when target tasks are easy, our models consistently outperform the BN models for all settings. 
In particular, for the hard domain adaptation to task ``A" on Table~\ref{tab:multi_office_result}, DSBN with source domain separation is considerably better than the {\it merged} case.
This results imply that DSBN has an advantage in multi-source domain adaptation tasks, too.
Note that the {\it seperate} case is not always better than the {\it merged} case without DSBN.

\begin{table}[t]
\centering
\caption{Classification accuracies (\%) in the multi-source scenario on Office-Home dataset using MSTN as a baseline. (ResNet-50)}
\label{tab:multi_office_home_result}
\vspace{0.1cm}
\scalebox{0.92}{
\begin{tabular}{lccc}
\toprule
\textit{Single} & Ar$\rightarrow$Rw & Cl$\rightarrow$Rw & Pr$\rightarrow$Rw\\
\hline
BN      & 76.4 & 69.3 & 76.9  \\
DSBN     & 75.4 & 70.4 & 77.7 \\
\toprule
\textit{Merged}        & \multicolumn{3}{c}{(Ar+Cl+Pr)$\rightarrow$Rw}\\
\hline
BN       & \multicolumn{3}{c}{81.2}\\
DSBN     & \multicolumn{3}{c}{82.3}\\
\hline
\textit{Separate}        & \multicolumn{3}{c}{Ar, Cl, Pr$\rightarrow$Rw}\\
\hline
BN       & \multicolumn{3}{c}{81.4}\\
DSBN     & \multicolumn{3}{c}{\bf{83.0}}\\
\bottomrule
\end{tabular}}
\end{table}

\begin{table*}[h]
\setlength\tabcolsep{5pt} \hspace{-0.3cm}
	\caption {Ablation results for the combination of batch normalization variations on VisDA-C validation dataset. (ResNet-101), where $\Delta$ means the accuracy gain by the second stage training with respect to the results from the first stage training only. } 
	\label{tab:visda_stage2}
	\vspace{0.1cm}
	\scalebox{0.9}{
		\begin{tabular}{cccccccclclclclclc|clclc}   
			\toprule
			Baseline&{Stage 1}&{Stage 2} &  {aero} & bicycle & bus & car & horse & knife & {motor} & person & plant & {skate} & train & truck & Avg & $\Delta$ \\ \hline
			\multirow{4}{*}{MSTN}&BN&BN	&81.2&48.7&77.8&75.3&90.2&4.6&\bf{94.4}&73.5&91.1&32.5&85.3&5.9&63.4 & -1.6\\
			&DSBN&BN 	&87.4&67.7&77.2&62.3&92.7&17.8&\bf{94.4}&74.5&\bf{92.5}&63.7&84.4&38.3&71.1 & -1.2\\
			&BN&DSBN	&92.8&70.3&\bf{78.4}&\bf{76.9}&93.3&14.9&92.7&\bf{81.7}&91.9&\bf{80.2}&85.0&20.3&73.2 & +7.2\\
			&DSBN&DSBN 	&\bf{94.7}&\bf{86.7}&76.0&{72.0}&\bf{95.2}&\bf{75.1}&87.9&81.3&91.1&68.9&\bf{88.3}&\bf{45.5}&\bf{80.2}& +7.9\\
			\hline
			\multirow{4}{*}{CPUA}&BN&BN	&88.9&43.9&69.8&\bf{68.8}&90.8&9.5&\bf{94.0}&46.0&\bf{92.1}&49.2&82.8&24.8&63.4&-2.8\\
			&DSBN&BN 	&89.3&53.6&77.9&54.2&91.7&16.1&93.7&74.3&91.6&58.4&79.6&52.4&69.4&-2.6\\
			&BN&DSBN	&91.1&66.0&79.3&\bf{68.8}&92.6&18.5&{91.2}&73.1&91.3&65.7&80.2&37.9&71.3&+4.7\\
			&DSBN&DSBN &\bf{93.0}&\bf{83.6}&\bf{79.9}&55.8&\bf{94.7}&\bf{20.6}&89.7&\bf{80.2}&{91.1}&\bf{80.8}&\bf{84.8}&\bf{59.7}&\bf{76.2}&+4.3\\
			\bottomrule
		\end{tabular}
	}
	\vspace{0.1cm}
\end{table*}

\subsection {Analysis}

\paragraph{Ablation Study} 
We perform ablative experiments on our framework to analyze the effect of DSBN in comparison to BN.
Table \ref{tab:visda_stage2} summarizes the ablation results on VisDA-C dataset using MSTN and CPUA as baseline architectures, where the last column in the table presents the accuracy gain by the second stage training with respect to the results from the first stage training only. 
We test several combinations of different training procedures for two stage training.
The results directly show that DSBN plays a crucial role on both the training procedures.
Another important point is that the second stage training with DSBN boosts performance additionally by large margins while the ordinary BN in the second stage is not helpful.
It implies that separating domain-specific information during training phase helps to get reliable pseudo-labels.
Note that, especially for hard classes, such a tendency is more pronounced.

\begin{figure}[t]
\centering
		\includegraphics[width=0.49\linewidth]{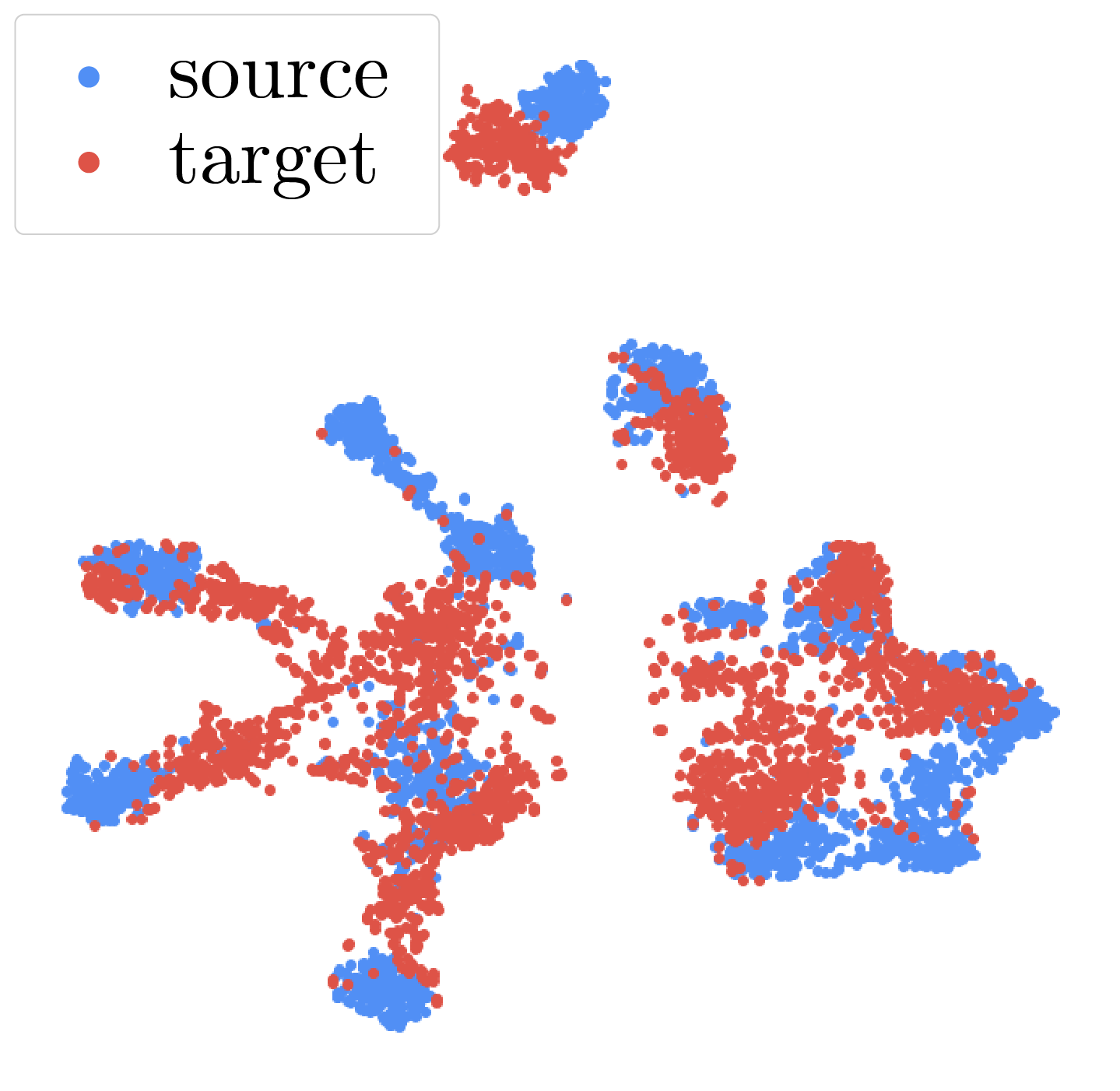}
		\includegraphics[width=0.49\linewidth]{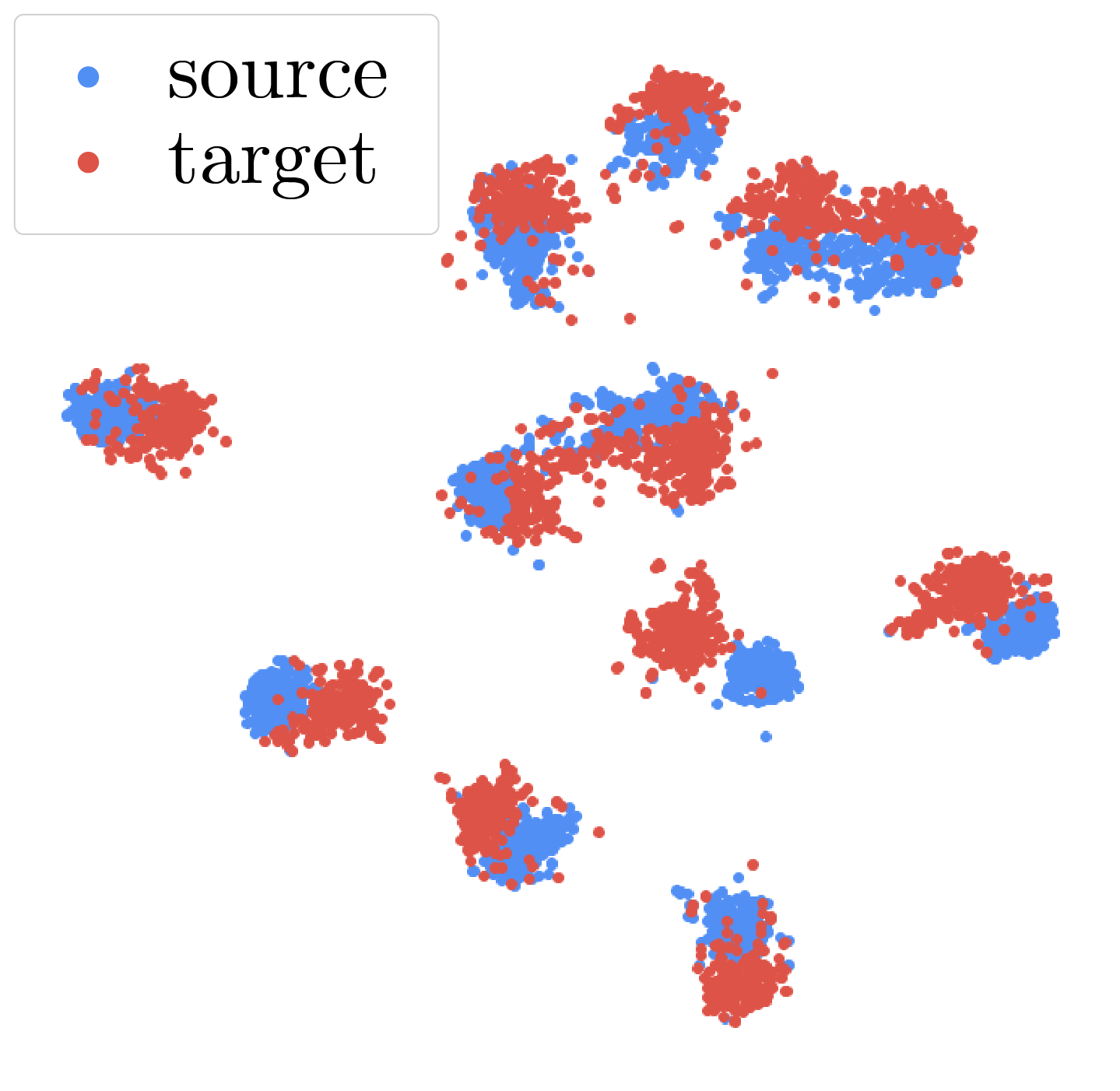}
		\vspace{0.1cm}
	\caption{t-SNE plots of the sample representations from ResNet-101 models trained with BN (left) and DSBN (right) using MSTN as a baseline algorithm on VisDA-C validation dataset. They illustrate that DSBN improves the consistency of representations across domains,}
	\label{fig:visda_refine_resnet101dsbn}
\end{figure}

\vspace{-0.2cm}
\paragraph{Feature Visualization} Figure~\ref{fig:visda_refine_resnet101dsbn} visualizes the instance embedding of BN (left) and DSBN (right) using MSTN as a baseline on VisDA-C dataset. 
We observe that the examples of two domains in the same class are aligned better by integrating DSBN, which implies that DSBN is effective to learn the domain-invariant representations.
 
\vspace{-0.2cm}
\paragraph{Iterative Learning} 
Our framework adopts the network obtained in the first stage as a pseudo labeler in the second stage, and the network learned in the second stage is stronger than the pseudo labeler.
Thus we can expect further performance improvement by applying the second stage learning procedure iteratively, where the pseudo-labels in the current iteration are given by the results in the preceding one.
To validate this idea, we evaluate the classification accuracy in each iteration on VisDA-C dataset using MSTN as a baseline algorithm.
As shown in Table \ref{tab:visda_iterative_result}, the iterative learning of the second stage gradually improves accuracy over iterations.

\begin{table}[t]
\centering
\caption{Classification accuracies (\%) of Iterative learning on VisDA-C dataset using MSTN as a baseline. (ResNet-101)}
\label{tab:visda_iterative_result}
\vspace{0.2cm}
\scalebox{0.9}{
\begin{tabular}{lccccc}
\toprule
 \multirow{2}{*}{Stage 1}&\multicolumn{4}{c}{Stage 2}\\
\cmidrule{2-5}
 & Iter 1 & Iter 2 & Iter 3 & Iter 4 \\
\hline
~~72.3& 80.2&81.4&82.2&{\bf 82.7}\\
\bottomrule
\end{tabular}}
\end{table}


\section{Conclusion}
\label{sec:conclusion}

We presented the domain-specific batch normalization for unsupervised domain adaptation.
The proposed framework has separate branches of batch normalization layers, one for each domain while sharing all other parameters across domains.
This idea is generically applicable to deep neural networks with batch normalization layers.
This framework with two stage training strategy is applied to two recent unsupervised domain adaptation algorithms, MSTN and CPUA, and demonstrates outstanding performance in both cases on the standard benchmark datasets.
We also present the capability of our framework to be extended to multi-source domain adaptation problems, and report significantly improved results compared to other methods.


\vspace{-0.2cm}
\paragraph{Acknowledgments} 
This work was partly supported by Kakao and Kakao Brain Corporation and Korean ICT R\&D program of the MSIP/IITP grant [2016-0-00563, 2017-0-01778].

{\small
\bibliographystyle{ieee}
\bibliography{main}

\begin{thebibliography}{10}\itemsep=-1pt

\bibitem{MMD}
Karsten~M. Borgwardt, Arthur Gretton, Malte~J. Rasch, Hans-Peter Kriegel,
  Bernhard Sch\"{o}lkopf, and Alex~J. Smola.
\newblock {Integrating Structured Biological Data by Kernel Maximum Mean
  Discrepancy}.
\newblock {\em Bioinformatics}, 22(14):e49--e57, July 2006.

\bibitem{DSN}
Konstantinos Bousmalis, George Trigeorgis, Nathan Silberman, Dilip Krishnan,
  and Dumitru Erhan.
\newblock {Domain Separation Networks}.
\newblock In {\em NIPS}, 2016.

\bibitem{DANN}
Yaroslav Ganin, Evgeniya Ustinova, Hana Ajakan, Pascal Germain, Hugo
  Larochelle, Fran{\c{c}}ois Laviolette, Mario Marchand, and Victor Lempitsky.
\newblock {Domain-Adversarial Training of Neural Networks}.
\newblock {\em JMLR}, 17(1):2096--2030, 2016.

\bibitem{hoffman2017cycada}
Judy Hoffman, Eric Tzeng, Taesung Park, Jun-Yan Zhu, Phillip Isola, Kate
  Saenko, Alexei~A. Efros, and Trevor Darrell.
\newblock {CyCADA: Cycle Consistent Adversarial Domain Adaptation}.
\newblock In {\em ICML}, 2018.

\bibitem{Batchnorm}
Sergey Ioffe and Christian Szegedy.
\newblock {Batch Normalization: Accelerating Deep Network Training by Reducing
  Internal Covariate Shift}.
\newblock In {\em ICML}, 2015.

\bibitem{Adamsolver}
Diederik~P. Kingma and Jimmy Ba.
\newblock {Adam: A Method for Stochastic Optimization}.
\newblock In {\em ICLR}, 2015.

\bibitem{adaptiveBN}
Yanghao Li, Naiyan Wang, Jianping Shi, Xiaodi Hou, and Jiaying Liu.
\newblock {Adaptive Batch Normalization for practical domain adaptation}.
\newblock {\em Pattern Recognition}, 80:109--117, 2018.

\bibitem{Mscoco}
Tsung-Yi Lin, Michael Maire, Serge Belongie, James Hays, Pietro Perona, Deva
  Ramanan, Piotr Doll{\'a}r, and C~Lawrence Zitnick.
\newblock {Microsoft COCO: Common Objects in Context}.
\newblock In {\em ECCV}, 2014.

\bibitem{DAN}
Mingsheng Long, Yue Cao, Jianmin Wang, and Michael~I Jordan.
\newblock {Learning Transferable Features with Deep Adaptation Networks}.
\newblock In {\em ICML}, 2015.

\bibitem{CDAN}
Mingsheng Long, Zhangjie Cao, Jianmin Wang, and Michael~I Jordan.
\newblock {Conditional Adversarial Domain Adaptation}.
\newblock In {\em NIPS}, 2018.

\bibitem{RTN}
Mingsheng Long, Han Zhu, Jianmin Wang, and Michael~I Jordan.
\newblock {Unsupervised Domain Adaptation with Residual Transfer Networks}.
\newblock In {\em NIPS}, 2016.

\bibitem{JAN}
Mingsheng Long, Han Zhu, Jianmin Wang, and Michael~I Jordan.
\newblock {Deep Transfer Learning with Joint Adaptation Networks}.
\newblock In {\em ICML}, 2017.

\bibitem{mancini2018boosting}
Massimiliano Mancini, Lorenzo Porzi, Samuel Rota~Bulò, Barbara Caputo, and
  Elisa Ricci.
\newblock {Boosting Domain Adaptation by Discovering Latent Domains}.
\newblock In {\em CVPR}, 2018.

\bibitem{CPUA}
Jeroen Manders, Elena Marchiori, and Twan van Laarhoven.
\newblock {Simple Domain Adaptation with Class Prediction Uncertainty
  Alignment}.
\newblock {\em arXiv preprint arXiv:1804.04448}, 2018.

\bibitem{autoDIAL}
Fabio Maria~Carlucci, Lorenzo Porzi, Barbara Caputo, Elisa Ricci, and Samuel
  Rota~Bulo.
\newblock {AutoDIAL: Automatic DomaIn Alignment Layers}.
\newblock In {\em ICCV}, 2017.

\bibitem{visda}
Xingchao Peng, Ben Usman, Neela Kaushik, Judy Hoffman, Dequan Wang, and Kate
  Saenko.
\newblock {VisDA: The Visual Domain Adaptation Challenge}, 2017.

\bibitem{office-31}
Kate Saenko, Brian Kulis, Mario Fritz, and Trevor Darrell.
\newblock {Adapting Visual Category Models to New Domains}.
\newblock In {\em ECCV}, 2010.

\bibitem{ATN}
Kuniaki Saito, Yoshitaka Ushiku, and Tatsuya Harada.
\newblock {Asymmetric Tri-training for Unsupervised Domain Adaptation}.
\newblock In {\em ICML}, 2017.

\bibitem{saito2017adversarial}
Kuniaki Saito, Yoshitaka Ushiku, Tatsuya Harada, and Kate Saenko.
\newblock {Adversarial Dropout Regularization}.
\newblock In {\em Proc. International Conference on Learning Representations
  (ICLR)}, 2018.

\bibitem{saito2017maximum}
Kuniaki Saito, Kohei Watanabe, Yoshitaka Ushiku, and Tatsuya Harada.
\newblock {Maximum Classifier Discrepancy for Unsupervised Domain Adaptation}.
\newblock In {\em CVPR}, 2018.

\bibitem{shen2017adversarial}
Jian Shen, Yanru Qu, Weinan Zhang, and Yong Yu.
\newblock {Wasserstein Distance Guided Representation Learning for Domain
  Adaptation}.
\newblock In {\em AAAI}, 2018.

\bibitem{CORAL}
Baochen Sun, Jiashi Feng, and Kate Saenko.
\newblock {Return of Frustratingly Easy Domain Adaptation}.
\newblock In {\em AAAI}, 2016.

\bibitem{DeepCORAL}
Baochen Sun and Kate Saenko.
\newblock {Deep CORAL: Correlation Alignment for Deep Domain Adaptation}.
\newblock In {\em ECCV Workshops}, 2016.

\bibitem{ADDA}
Eric Tzeng, Judy Hoffman, Kate Saenko, and Trevor Darrell.
\newblock {Adversarial Discriminative Domain Adaptation}.
\newblock In {\em CVPR}, 2017.

\bibitem{DDC}
Eric Tzeng, Judy Hoffman, Ning Zhang, Kate Saenko, and Trevor Darrell.
\newblock {Deep Domain Confusion: Maximizing for Domain Invariance}.
\newblock {\em CoRR}, abs/1412.3474, 2014.

\bibitem{office-home}
Hemanth Venkateswara, Jose Eusebio, Shayok Chakraborty, and Sethuraman
  Panchanathan.
\newblock {Deep Hashing Network for Unsupervised Domain Adaptation}.
\newblock In {\em CVPR}, 2017.

\bibitem{MSTN}
Shaoan Xie, Zibin Zheng, Liang Chen, and Chuan Chen.
\newblock {Learning Semantic Representations for Unsupervised Domain
  Adaptation}.
\newblock In {\em ICML}, 2018.

\bibitem{WMMD}
Hongliang Yan, Yukang Ding, Peihua Li, Qilong Wang, Yong Xu, and Wangmeng Zuo.
\newblock {Mind the Class Weight Bias: Weighted Maximum Mean Discrepancy for
  Unsupervised Domain Adaptation}.
\newblock In {\em CVPR}, 2017.

\bibitem{CMD}
Werner Zellinger, Thomas Grubinger, Edwin Lughofer, Thomas Natschl{\"a}ger, and
  Susanne Saminger-Platz.
\newblock {Central Moment Discrepancy (CMD) for Domain-Invariant Representation
  Learning}.
\newblock In {\em ICLR}, 2017.

\bibitem{iCAN}
Weichen Zhang, Wanli Ouyang, Wen Li, and Dong Xu.
\newblock {Collaborative and Adversarial Network for Unsupervised domain
  adaptation}.
\newblock In {\em CVPR}, 2018.

\end{thebibliography}
}

\end{document}